\documentclass[journal]{IEEEtran}
\usepackage{cite}
\usepackage{caption}  % Add this package for caption formatting
\usepackage{url}
\usepackage{subfig}
\ifCLASSINFOpdf
\usepackage[pdftex]{graphicx}
\else
\fi
\begin{document}
\title{Teaching Wav2Vec2 the Language of the Brain}
\author{Tobias Fiedler\textsuperscript{1,*},
        Leon Hermann\textsuperscript{1,*},
        Florian Müller\textsuperscript{1}, 
        Sarel Cohen\textsuperscript{2},
        Peter Chin\textsuperscript{3},
        Tobias Friedrich\textsuperscript{1},
        Eilon Vaadia\textsuperscript{4}\\
        
        \vspace{5pt}
        \textsuperscript{1}Hasso Plattner Institute, Digital Engineering Faculty, University of Potsdam,
        \texttt{\{tobias.fiedler,leon.hermann,florian.mueller\}@student.hpi.de}, \texttt{tobias.friedrich@hpi.de}\\
        \textsuperscript{2}The Academic College of Tel Aviv-Yaffo, Israel, \texttt{sarelco@mta.ac.il}\\
        \textsuperscript{3}Dartmouth College, Hanover, NH 03755, \texttt{peter.chin@dartmouth.edu}\\
        \textsuperscript{4}The Edmond and Lily Safra Center for Brain Sciences at The Hebrew University of Jerusalem, Israel, \texttt{eilon.vaadia@elsc.huji.ac.il}\\
        \textsuperscript{*}Equal contribution
}
\markboth{Teaching Wav2Vec2 the Language of the Brain}%
{Shell \MakeLowercase{\textit{et al.}}: Bare Demo of IEEEtran.cls for IEEE Journals}

\maketitle
\begin{abstract}
The decoding of continuously spoken speech from neuronal activity has the potential to become an important clinical solution for paralyzed patients. Deep Learning Brain Computer Interfaces (BCIs) have recently successfully mapped neuronal activity to text contents in subjects who attempted to formulate speech. However, only small BCI datasets are available. 
In contrast, labeled data and pre-trained models for the closely related task of speech recognition from audio are widely available.
One such model is Wav2Vec2 which has been trained in a self-supervised fashion to create meaningful representations of speech audio data. 
In this study, we show that patterns learned by Wav2Vec2 are transferable to brain data.
Specifically, we replace its audio feature extractor with an untrained Brain Feature Extractor (BFE) model. We then execute full fine-tuning with pre-trained weights for Wav2Vec2, training “from scratch” without pre-trained weights as well as freezing a pre-trained Wav2Vec2 and training only the BFE each for 45 different BFE architectures.  
Across these experiments, the best run is from full fine-tuning with pre-trained weights, achieving a Character Error Rate (CER) of 18.54\%, outperforming the best training from scratch run by 20.46 and that of frozen Wav2Vec2 training by 15.92 percentage points. 
% WER VERSION: Across these experiments, the best run is from Full Fine-Tuning with pre-trained weights, achieving a Word Error Rate (WER) of 30.97\% without LM decoding, outperforming the best run of training from scratch by 35.24 and that of BFE only training by 22.32 percentage points. 
These results indicate that knowledge transfer from audio speech recognition to brain decoding is possible and significantly improves brain decoding performance for the same architectures. Related source code is available at \url{https://github.com/tfiedlerdev/Wav2Vec2ForBrain}.

\end{abstract}
\begin{IEEEkeywords}
Neural Signals, Deep Learning, Cross-modal Transfer Learning
\end{IEEEkeywords}

\section{Introduction}
%MOTIVATION%
\IEEEPARstart{A}{s} a social being, the ability to communicate is essential to the mental health of human individuals.  However, some people have lost theirs e.g. through a brainstem stroke or ALS, limiting their ability to participate socially. 
The new developments of brain-computer interfaces (BCIs) take the challenge to help these patients.

The method involves recordings of neuronal spiking activity via hundreds of micro-electrodes implanted directly in areas of interest in the participant's brain.
By recording this neuronal activity as the participant attempts to speak a given sentence, a dataset of brain data and corresponding attempted speech content is created. 
The computational challenge is to map the neuronal activity time series data to the corresponding text, essentially discovering patterns within the brain data and mapping them to attempted spoken characters, words, or phonemes, to produce a plain English transcription from unseen brain activity.
What makes this task especially challenging is the small amount of available data.
In contrast, data and pre-trained models for the closely related task of audible speech recognition are widely available.

One such model is Wav2Vec2 which has been pre-trained in a self-supervised fashion to create meaningful representations of speech audio data and finetuned to transcribe speech based on these representations.
In this study, we show that patterns learned by Wav2Vec2 for the task of speech recognition are transferable to the task of neuronal spiking activity based on attempted speech decoding.
Specifically, we replace the audio feature extractor of Wav2Vec2 with an untrained Brain Feature Extractor (BFE) model. We then execute "Full Fine-Tuning" with pre-trained weights for Wav2Vec2, "Training from Scratch” without pre-trained weights as well as "Frozen Wav2Vec2" training with pre-trained and locked weights for Wav2Vec2 each for 45 different BFE architectures.

Across all experiments, the best run is from Full Fine-Tuning with pre-trained weights, achieving a Word Error Rate (WER) of 30.97\% and a Character Error Rate (CER) of 18.54\% without Language Model (LM) assisted decoding, while the best runs of Training from Scratch and Frozen Wav2Vec2 training achieve significantly worse WERs of 66.21\% (39.01\% CER) and 53.29\% (32.46\% CER) respectively. 
These results show that knowledge transfer from audio speech recognition to neuronal activity decoding works.
Additionally, the Frozen Wav2Vec2 training runs having a CER better than a random predictor show that neuronal spiking data can be transformed into a representation understood by models trained on different input domains such as audio signals. 
\\
\section{Related Work}
Recently, there have been major advances in the field of BCIs. This report focuses on related work on decoding spoken sentences from brain activity.

This project was initially set up to participate in a deep learning challenge called brain-to-text published by Willett et al.\cite{b2t_original}. As such, the model architectures explained in this paper were trained on the dataset they collected. Willet et al. implemented a GRU\cite{gru} model architecture to predict spoken phonemes from brain activity and added a language model in post processing with a resulting WER of 23.8\% on a 125,000-word vocabulary. However, our experiments are executed on a lower training sample count since the provided dataset did not supply transcriptions of data samples in the competition holdout set. Thus, we created a test set  from the training samples, reducing the size of the training partition.

Benster et al.\cite{benster2024cross} developed a multimodal architecture that can be applied to arbitrary pairs of speech modalities. On the brain-to-text dataset, they were able to improve the previous best contribution from 9.8\% WER to 8.9\%. Similarly to Willet et al. experiments, they used the complete training set without partitioning it further. 

At the point of submission of the study, we are aware of one competitor of the brain-to-text challenge which achieves a WER of 5.81\%.

In this work, we explored the ability of the state-of-the-art speech transcription transformer\cite{transformer} model "Wav2Vec2"\cite{w2v2} to apply its learned patterns to the given task of converting brain recordings to text. To our best knowledge, there are no related works that attempted to apply a pre-trained speech transcription transformer like Wav2Vec2 to the input domain of brain recordings. However, Wu et al.\cite{speechLlama} have successfully adapted the pre-trained LLM called "Llama" to audio signals, roughly resembling our approach on different domains.  
\section{Methods}
\subsection{Dataset}
We used the brain-to-text dataset\cite{b2t_original} as stated in the related work section. The creators collected the data for five days of vocalized speech and three days of silent speech. In total, 10,850 sentences were recorded. The brain activity of a singular patient was measured with four microelectrode arrays. Two of these were implanted in the ventral premotor cortex and the remaining two were in area 44, which is part of Broca's area. The work of the brain-to-text researchers suggested that area 44 contained little information for speech classification so we also excluded that data from training.

The data samples were recorded in blocks. Each day has a fixed amount of blocks. These blocks have already been split by the Brain-to-Text team into a training, test, and competition holdout partition. This project uses the given test set as a validation set during training. Since the targets of the competition holdout set are not supplied, we chose to create our test set by taking the first block of each day of the train set as test data. With this, we decrease the data available for training from 8,800 recorded sentences to 7,920 and therefore take a worse model performance into account.

Each block contains a fixed amount of sentences and the corresponding brain activity recordings. The recorded microelectrode data is already available in a pre-processed form. For each 20 ms window, the count of threshold crossings and the mean squared spike power are given for each microelectrode.
\subsection{Wav2Vec2 for Brain Data}
The original Wav2Vec2 model \cite{w2v2} consists of a CNN model for audio feature extraction, an encoder-only transformer module for contextualization of those features, and a fully connected language modeling head for character classification.
The extracted audio features are pre-trained to resemble speech units that correspond to phonemes. \\
\begin{figure}[!htb]
    \centering
    \includegraphics[width=0.9\linewidth]{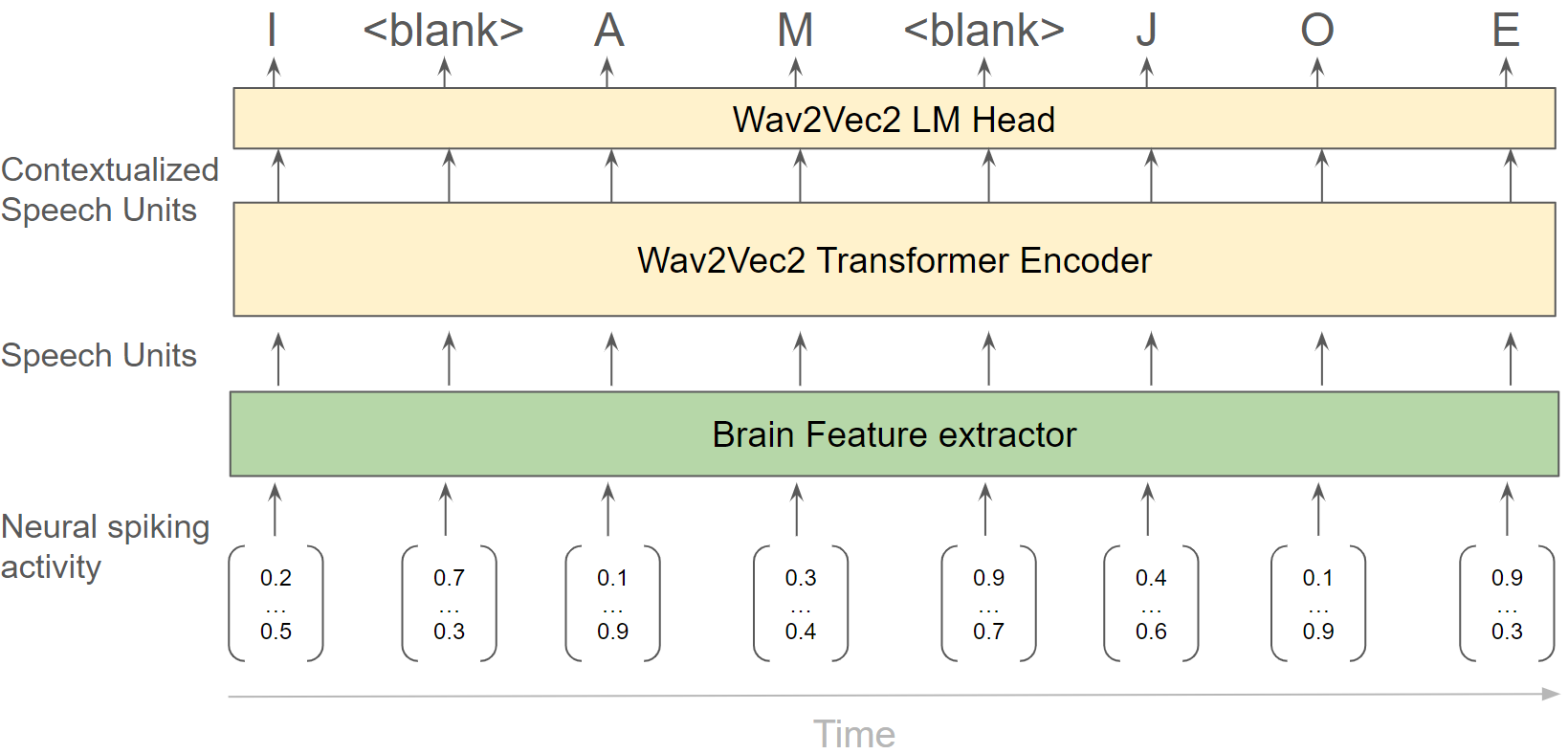}
    \caption{"Wav2Vec2 for CTC loss" architecture adapted to brain data. Here, the Wav2Vec2 feature extractor is replaced by our BFE, which is a GRU model followed by a fully connected projection network.}
    \label{fig:bfe_w2v_architecture}
\end{figure}

To adapt Wav2Vec2 to brain data, we exchange its audio feature extraction model with a bidirectional GRU model followed by fully connected layers to project from the GRU output shape to the shape expected by the Wav2Vec2 transformer.
Willet et al. \cite{b2t_original} have shown GRUs to be able to extract phoneme sequences from brain data, indicating that they could also produce speech units as expected by the Wav2Vec2 transformer module. The term "speech units" was introduced in the Wav2Vec2 paper\cite{w2v2}, describing the latent representations of the corresponding audio signal as produced by the Wav2Vec2 audio feature extractor. They were found to correspond to phonemes after the self-supervised pre-training.

Therefore, we replicate the preprocessing as done by Willet et al., including a feature and recording-block-wise z-scoring, a fully connected layer with uniquely learned parameters for each recording day, and rolling feature adaptation. A depiction of the adapted architecture can be found in figure~\ref{fig:bfe_w2v_architecture}.

\subsection{Experimental Setups}
To test if knowledge from Wav2Vec2 pretraining is transferable to the task of brain decoding, we compare three different experiment setups:

\begin{enumerate}
    \item "Full Fine-Tuning": Training of all modules with pre-trained weights for Wav2Vec2 Transformer and Language Modelling Head
    \item "Frozen Wav2Vec2": Freezing the pre-trained Wav2Vec2 parameters and training only the BFE
    \item "Training from Scratch": Training of all modules without pre-trained weights for the Wav2Vec2 modules 
\end{enumerate}

The pre-trained weights for Wav2Vec2 are loaded from the checkpoint at \url{https://huggingface.co/facebook/wav2vec2-base-960h}. The optimization objective of all experiments is to minimize the Connectionist Temporal Classification\cite{ctc_loss} (CTC) loss over the logits produced by the Wav2Vec2 LM Head.

\begin{figure}[!htb]
    \centering
    \includegraphics[width=1\linewidth]{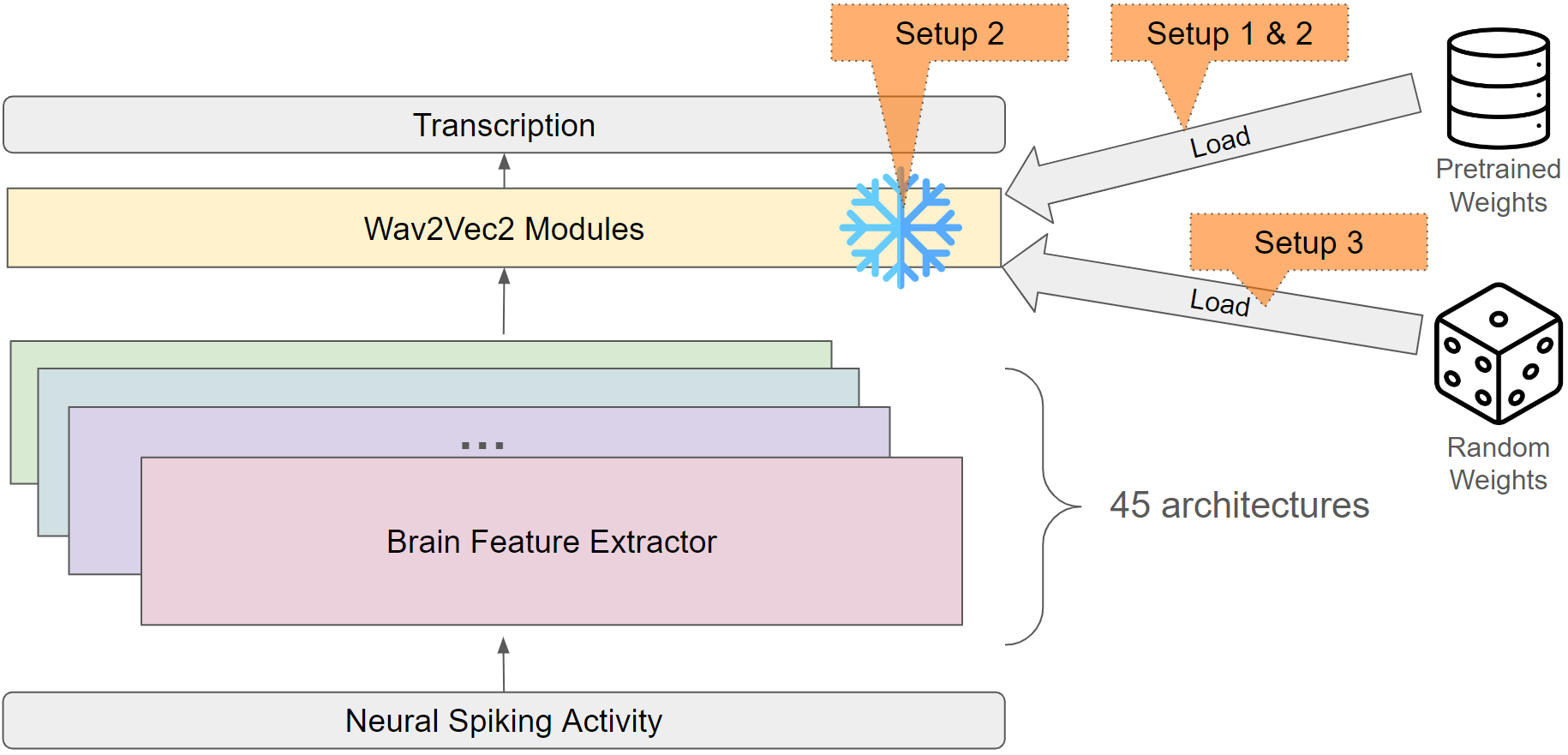}
    \caption{Conceptual depiction of experiment setups. Setup 1 and 2 load pre-trained weights for the Wav2Vec2 model, while Setup 3 loads random weights. Setup 2 freezes the Wav2Vec2 module parameters. In all setups, training runs with the same 45 BFE architectures are executed.}
    \label{fig:experiment_setup}
\end{figure}

For each of the three setups, we test the same 45 BFE architectures. Specifically, we test three GRU depths (1, 3 and 5 layers) across three GRU hidden sizes (256, 512 and 1024) and five fully connected depths and widths ($[]$ i.e. only project from GRU output shape to Wav2Vec2 transformer input shape, $[128]$ i.e. one hidden layer of width 128, $[256]$, $[512]$ and $[512, 128]$ i.e. two hidden layers of width 512 and 128 respectively).
The two setups with pre-trained Wav2Vec2 parameters (Full Fine-Tuning and Frozen Wav2Vec2) are trained with a Wav2Vec2 module-specific learning rate of $0.00001$ after ten epochs, linearly increasing from $0$, starting from the seventh epoch. Thus, we make sure its weights are not destroyed while the BFE outputs and therefore the Wav2Vec2 gradients are still random.
Apart from these, we keep all hyperparameters equal: Each run is executed for 100 epochs with an early stopping patience of ten epochs, meaning a run is stopped early if there is no improvement in the CTC-loss on the validation set for ten consecutive epochs.
The general learning rate is set to $0.0001$.
After the training, the model checkpoint with the best validation error rate of all training epochs is evaluated on the test set.
A complete list of the fixed hyperparameters can be found in our repository at \url{https://github.com/tfiedlerdev/Wav2Vec2ForBrain/tree/main/sweeps}.
\section{Results}
% Sweeps
% Visualization with models above 90% word error rate

% Observe the increase in performance for pre-trained and finetuning
%In the following section, we compare the performance of the experiment setups via an empirical analysis and then perform an exploratory analysis of the latent representations produced by the Brain Feature Extractor and the original Wav2Vec2 Audio Feature Extractor.%
\subsection{Comparison of Experiment Setups}
\begin{figure}[!htb]
    \centering
    \includegraphics[width=0.9\linewidth]{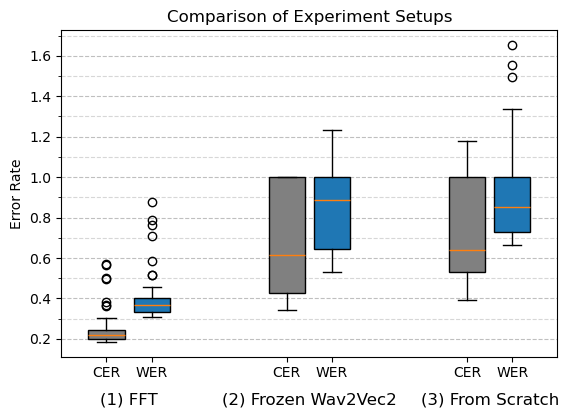}
    \caption{Comparison of the CER and WER achieved on the test set by the runs of the three experiment setups: (1) Full Fine-Tuning, (2) Frozen Wav2Vec2 training, and (3) Training from Scratch.}
   \label{fig:sweep-results}
\end{figure}

A visual comparison of the results of the experimental setups can be seen in Figure \ref{fig:sweep-results}. It clearly shows that Full Fine-Tuning with pre-trained Wav2Vec2 modules outperforms the other setups in most cases. Furthermore, it highlights that some Brain Feature Extractor architectures within Frozen Wav2Vec2 perform better than all architectures within training from scratch, indicating knowledge transfer from pretraining is beneficial if the right architecture is chosen. 
Specifically, within the Full Fine-Tuning setup, 40 out of 45 Brain Feature Extractor architectures achieved lower WERs than the best run out of both of the other setups.
Statistically, the difference in WER is also significant: The WER of the Full Fine-Tuning runs are significantly lower than those of Frozen Wav2Vec2 (one-sided Wilcoxon signed-rank test, $p < 3\cdot 10^{-14}$) and those of Training from Scratch ($p < 9\cdot 10^{-14}$). Also, the WER of Frozen Wav2Vec2 runs are significantly smaller than those of Training from Scratch ($p=0.0562$), even though it is less certain here.
When comparing the Frozen Wav2Vec2 runs to the Training from Scratch runs, 13 out of its 45 runs still achieve lower WERs than the best run of Training from Scratch.
\begin{table}[!htb]
    \centering
    \begin{tabular}{|p{3.5cm}||p{0.8cm}|p{0.8cm}|p{1.4cm}|}
        \hline
        \textbf{Setup} & \textbf{CER} & \textbf{WER} & \textbf{WER (LM)} \\
        \hline
        (1) Full Fine-Tuning & 18.54\% & 30.97\% & 28.61\% \\
        (2) Frozen Wav2Vec2 & 34.46\% & 53.29\% & 52.41\% \\
        (3) Training from Scratch & 39.01\% & 66.21\% & 58.73\% \\
        ($*$) FFT \& W2V2Conformer & 15.72\% & 26.67\% & 26.81\% \\
        \hline
    \end{tabular}
    \vspace{0.2cm}
    \caption{The runs with the best WER within each experimental setup and their respective Character Error Rate and Word Error Rate when evaluated on the test partition. Language Model (LM) decoding is done via 3-gram LM assisted Beam Search for CTC-logit decoding.\\
    $*$ Full Fine Tuning with pre-trained Wav2Vec2Conformer  }
    \label{tab:best-runs}
\end{table}

In numbers, the best run of Full Fine-Tuning achieved a WER of 30.97\% and a CER of 18.54\%, the one of Frozen Wav2Vec2 53.29\% and 34.46\% and Training from Scratch 66.21\% and 39.01\% as can be seen in Table~\ref{tab:best-runs}. With LM assisted logit decoding, we observe a small improvement in WER for all of these architectures. The only configuration outperforming the best regular Full Fine-Tuning run is Full Fine-Tuning with a pre-trained Wav2Vec2Conformer\footnote{Pre-trained Wav2Vec2Conformer checkpoint from \url{https://huggingface.co/facebook/wav2vec2-conformer-rope-large-960h-ft}} model, which was not included in the regular experiment setup due to computational restrictions. It uses a conformer network instead of a transformer encoder. Here, we tested only a subset of BFE architectures and achieved a WER of 26.67\% and CER of 15.72\%.

Next, we compare how many BFE architectures learned any general patterns within each setup. For that, we interpret any CER equal to or greater than 100\% as the model not having learned useful patterns. Within the Full Fine-Tuning setup, all BFE architectures achieved a test CER of at most 56.96\%, indicating that this setup reliably learns useful patterns.
In contrast, 14 out of 45 Frozen Wav2Vec2 runs and 13 out of 45 Training from Scratch runs had a CER equal to or higher than 100\%, showing that the chosen BFE Architecture is of high importance regarding general training success in these setups.\\
When directly comparing each BFE architecture across the three setups, in 44 out of 45 cases Full Fine-Tuning outperforms the same architecture in the other setups in terms of WER. Only for one architecture, Training from Scratch outperforms the same architecture in both other setups. For this architecture, Full Fine-Tuning has the worst WER across all architectures, specifically 87.46\%, while Training from Scratch achieves 69.31\% and Frozen Wav2Vec2 100\%, indicating that this specific architecture cannot generate features useful for a pre-trained Wav2Vec2.
When comparing the architectures directly between Frozen Wav2Vec2 and Training from Scratch, Frozen Wav2Vec2 has a lower WER on 25 out of 31 architectures that were capable of producing features usable by the frozen Wav2Vec2 modules in the first place, i.e. those where Frozen Wav2Vec2 has a CER lower than 100\%. This indicates that given a BFE architecture capable of producing sensible features for Wav2Vec2, pre-trained Wav2Vec2 modules outperform non-pre-trained ones in most cases.

\subsection{Latent Analysis}
% Data analysis

We compare the features extracted by the pre-trained Wav2Vec2 audio feature extractor to the features extracted by the BFE, both before and after contextualization through the Wav2Vec2 transformer.
For that, we synthesize the speech for the transcriptions within the test set via the ElevenLabs tool\footnote{ElevenLabs tool for speech synthesis \url{https://elevenlabs.io/}}. To make the audio and brain features comparable, we perform the analysis on the best BFE out of the Frozen Wav2Vec2 setup. i.e. that with the best test CER. For all other setups, the features of the two domains would not align as the Wav2Vec2 modules processing them are trained and therefore changed, hence requiring different features than generated by the pre-trained Wav2Vec2 audio feature extractor.

\begin{figure}[!t]
\centering
\subfloat[Before Wav2Vec2 transformer]{\includegraphics[width=0.49\columnwidth]{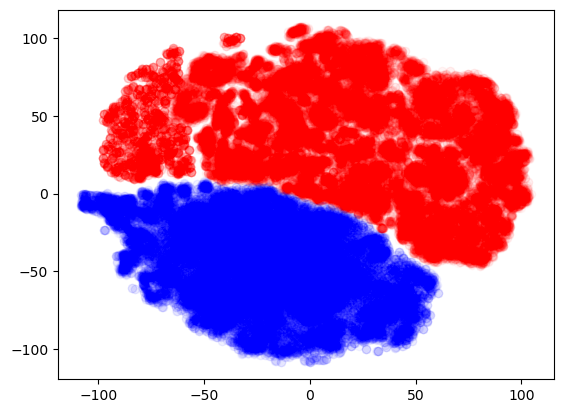}%
\label{fig:latent_pre_w2v}}
\hfil
\subfloat[After Wav2Vec2 transformer]{\includegraphics[width=0.49\columnwidth]{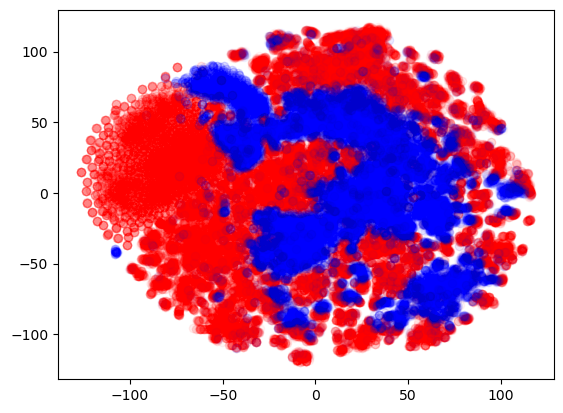}%
\label{fig:latent_post_w2v}}
\caption{Features extracted by BFE (blue) and pre-trained Wav2Vec2 Audio Feature Extractor (red). 
Both depicted before and after being passed through the same pre-trained Wav2Vec2 transformer module. Reduced in dimension via t-SNE.}
\label{fig:latent_w2v_combined}
\end{figure}
%\begin{figure}
%    \centering
%    \includegraphics[width=1\linewidth]{assets/pre_w2v.png}
%    \caption{Features extracted by Brain Feature Extractor and pre-trained Wav2Vec2 Audio Feature Extractor before being passed into the Wav2Vec2 transformer, reduced in dimension via t-SNE.}
%    \label{fig:latent_pre_w2v}
%\end{figure}
The dimensionality-reduced brain and audio features are separated as can be seen in Figure~\ref{fig:latent_pre_w2v}, indicating a strong difference in distribution between extracted audio and brain features. While this specific BFE architecture achieved a better-than-random CER of 34.46\% together with the frozen Wav2Vec2 modules, the BFE does seem to fail in resembling the feature distribution as expected by the Wav2Vec2 transformer.
A possible cause of this is the difference in architecture: while the audio feature extractor uses a stateless CNN, the BFE utilizes a stateful GRU. The CNN can only incorporate the local context of the input data, while the bidirectional GRU generates its outputs based on the whole input sequence. However, this makes it even more noteworthy that the pre-trained Wav2Vec2 transformer is capable of dealing with these differences and is still able to produce sensible outputs.

%\begin{figure}[!htb]
%    \centering
%    \includegraphics[width=1\linewidth]{assets/post_w2v.png}
%    \caption{Features extracted by Brain Feature Extractor and pre-trained Wav2Vec2 Audio Feature Extractor after being passed into the Wav2Vec2 transformer, i.e. being contextualized, reduced in dimension via t-SNE.}
%    \label{fig:latent_post_w2v}
%\end{figure}
When looking at the dimensionality-reduced features after being contextualized by the pre-trained Wav2Vec2 transformer (see Figure~\ref{fig:latent_post_w2v}), one can see a weaker separation with noticeable overlap. This indicates that the pre-trained Wav2Vec2 transformer is at least partially capable of aligning the differently distributed brain features with the learned audio feature distribution.

% text corresponding to other clusterings put into https://docs.google.com/document/d/1R8hkFBUA2LMtxA9UANlvAl_nOBg3TL_EHG_IkvKSf1Q/edit

\section{Discussion \& Conclusion}
% Discuss sweep observations
% - Statistically significant findings?
% - Further improvements possible?
%   - Pretraining on large amounts of brain recordings
The experiment results show a significant increase in performance when using a pre-trained instead of an untrained Wav2Vec2 model. However, some aspects need to be considered. %Firstly, our experiment training "from scratch" had access to a small dataset. Considering that transformers are known to require a large amount of data to train, the inferiority of this approach is expected, highlighting the need for larger datasets or knowledge transfer. 
Firstly, the experiment applying a pre-trained Wav2Vec2 conformer model yielded promising results. Since this architecture could not be investigated in detail, further research needs to be done to explore its full potential.
Secondly, when comparing the latent distributions of features extracted from brain to those of audio signals, a clear separation can be observed.  However, the brain signal features are still interpretable by the pre-trained transformer module, as shown by our experiment results and by a decrease in separation when jointly visualizing the audio and brain features after they were contextualized by the transformer. 
To mitigate this separation, other BFE architectures than the GRU should be explored, e.g. the recently introduced Selective State Space model called Mamba\cite{mamba}. 
Lastly, more sophisticated CTC logit decoding approaches, like LLM-assisted beam search decoding should be investigated to improve performance further.
% Discuss data analysis
% - Similar distribution?
% - Discuss the difference between before and after wav2vec

% Conclusion of Discussion: Does pretraining on speech-to-text tasks transfer to brain-to-text tasks? Specifically Wav2Vec2
To conclude, we have shown that knowledge transfer from the speech recognition task to that of brain decoding does work, and significantly increases performance when comparing the same model architecture in a pre-trained and untrained state. 
As small sample sizes are a common limitation for BCI datasets, knowledge transfer is crucial especially for transformer architectures that require large datasets to learn general patterns. Further investigation into this approach may yield performance improvements beyond the current state of the art of neuronal activity decoding, further extending the real world applicability of BCIs as clinical solutions.

%As long as this limitation is not removed, input domain adaptation of pre-trained transformer models to the domain of brain data needs to be investigated further. We believe that this approach holds a lot more potential than we were already able to explore.

%With this, the limitation of small BCI datasets can be mitigated, which is especially useful for Transformer architectures that require large datasets to learn general patterns.
%As small sample sizes are a common limitation for brain-to-text prediction, applying speech-to-text pre-training to this task can compensate for the lack of brain recording data for self-supervised training of the transformer. 

%However, the most desirable solution would be to collect an extensive dataset of brain recordings of multiple patients to enable a pretraining of a Wav2Vec2-like transformer on brain recordings directly. We expect this approach to yield the best prediction quality once a dataset of adequate sample count is available.

\appendices

\ifCLASSOPTIONcaptionsoff
  \newpage
\fi



%\bibliographystyle{IEEEtran}
\bibliographystyle{IEEEtran}
\bibliography{bibtex/bib/ICASSP}
\end{document}